%% file: New_IEEEtran_how-to.tex
\pdfoutput=1
\documentclass[lettersize,journal]{IEEEtran}
\usepackage{amsmath,amsfonts}
\usepackage{algorithmic}
\usepackage{array}
\usepackage[caption=false,font=footnotesize,labelfont=rm,textfont=rm]{subfig}
\usepackage{textcomp}
\usepackage{stfloats}
\usepackage{url}
\usepackage{verbatim}
\usepackage{graphicx}

\usepackage[table,xcdraw]{xcolor}
\definecolor{GREEN}{RGB}{175, 211, 153}
\definecolor{RED}{RGB}{255, 190, 190}
\definecolor{BLUE}{RGB}{28, 152, 214}

\usepackage{multirow}
\usepackage[ruled]{algorithm2e}
\usepackage{setspace}
\usepackage{makecell}  % 表格内换行
\usepackage{amsmath}
\usepackage{amssymb}  % 特殊符号
\usepackage{subfig}
\usepackage{float}
\usepackage{booktabs}
\usepackage{cite}
\usepackage{lettrine}

\usepackage{balance}

\begin{document}
\title{Tackling the Non-IID Issue in Heterogeneous Federated Learning by Gradient Harmonization}
\author{Xinyu Zhang, Weiyu Sun, Ying Chen
\thanks{
This work was mainly supported by the Fundamental Research Funds for the Central Universities under Grant 2023300311, 0210-14380209.

Xinyu Zhang, Weiyu Sun, and Ying Chen are with the School of Electronic Science and Engineering, Nanjing University, Nanjing 210023, China. E-mail: \{xinyuzhang, weiyusun\}@smail.nju.edu.cn, \{yingchen\}@nju.edu.cn.}}

\markboth{Journal of \LaTeX\ Class Files,~Vol.~18, No.~9, September~2020}%
{How to Use the IEEEtran \LaTeX \ Templates}

\maketitle

\begin{abstract}
    Federated learning (FL) is a privacy-preserving paradigm for collaboratively training a global model from decentralized clients. 
    However, the performance of FL is hindered by non-independent and identically distributed (non-IID) data and device heterogeneity. 
    In this letter, we revisit this key challenge through the lens of gradient conflicts on the server side. 
    Specifically, we first investigate the gradient conflict phenomenon among multiple clients and reveal that stronger heterogeneity leads to more severe gradient conflicts. 
    To tackle this issue, we propose FedGH, a simple yet effective method that mitigates local drifts through Gradient Harmonization. 
    This technique projects one gradient vector onto the orthogonal plane of the other within conflicting client pairs. 
    Extensive experiments demonstrate that FedGH consistently enhances multiple state-of-the-art FL baselines across diverse benchmarks and non-IID scenarios. 
    Moreover, FedGH yields more significant improvements in scenarios with stronger heterogeneity. As a plug-and-play module, 
    FedGH can seamlessly integrate into any FL framework without requiring hyperparameter tuning.
\end{abstract}

\begin{IEEEkeywords}
    Federated learning, non-IID issue, gradient conflict, gradient harmonization, robust server aggregation
\end{IEEEkeywords}
\vspace{-0.3cm}

\input{sections/1_intro.tex}
\input{sections/3_methodology.tex}
\input{sections/4_experiments.tex}

\section{Conclusion}
The non-IID issue has become a key challenge as federated learning research advances. In this letter, we first investigate the gradient conflict phenomenon among multiple clients in FL. Then, we propose a simple yet effective method called FedGH to tackle the non-IID issue through gradient harmonization. FedGH can seamlessly integrate into any FL framework without hyperparameters tuning. Extensive experiments demonstrate that FedGH consistently improves multiple state-of-the-art baselines across different degrees of heterogeneity, indicating its potential for real-world applications.

% 放在\end{document}前
\bibliographystyle{IEEEtran}
\bibliography{IEEE}

\end{document}

%% file: sections/1_intro.tex
\section{Introduction}
\lettrine{U}{NDER} the prevalence of large deep-learning models \cite{GPT, ViT}, the demand for massive data overgrows. 
However, these datasets are usually dispersed across heterogeneous devices (e.g., mobile devices, personal computers, company servers, etc.), 
and conventional centralized training is constrained due to privacy concerns \cite{DP, kaissis2020secure, Inverting_gradients}. In this scenario, federated learning (FL) \cite{FedAvg}, 
a privacy-preserving distributed training strategy, has emerged as a promising alternative. 
FL aims to collaboratively learn a global model based on multiple clients without data sharing. 
Specifically, each communication round in FL can be divided into three steps \cite{FedAvg}. First, clients download the global model to update the local one. 
Next, each client performs local training and uploads the local gradient. 
Finally, the server aggregates local updates to generate the global model. 
Driven by the growing need for privacy protection and decentralized training, FL has been applied widely, including but not limited to face recognition \cite{face}, semantic segmentation \cite{FedSeg}, 
medical image analysis \cite{Harmofl}, and wireless communication \cite{wireless}.

\begin{figure}
    \vspace{-0.3cm}
    \centering
    \subfloat[]{\includegraphics[width=0.16\textwidth]{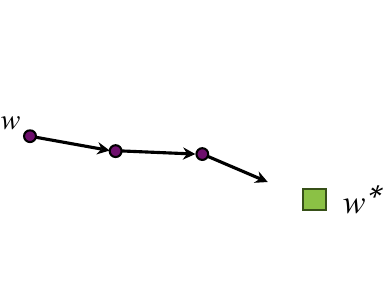}\label{Centralized training}}
    \subfloat[]{\includegraphics[width=0.16\textwidth]{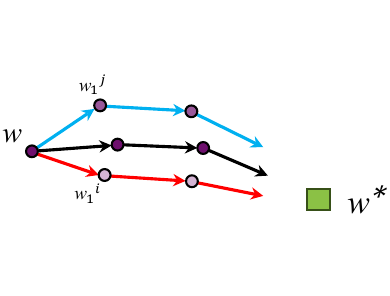}\label{Homogenous FL}}
    \subfloat[]{\includegraphics[width=0.17\textwidth]{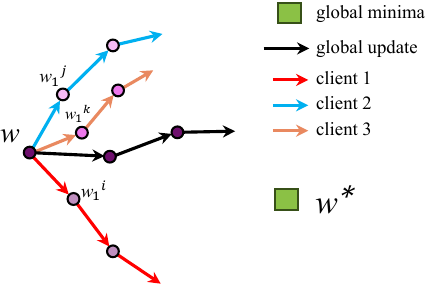}\label{Heterogeneous FL}}
    \caption{Illustration of optimization directions from (a) centralized training, (b) homogenous FL, and (c) heterogeneous FL with gradient conflicts.}
    \label{local drift}
    \vspace{-15pt}
\end{figure}

Despite its advanced characteristics, 
FL suffers from local drifts \cite{scaffold} across multiple clients due to the non-independent and identically distributed (non-IID) issue, 
as illustrated in Fig. \ref{local drift}. 
In a homogenous scenario (Fig. \ref{Homogenous FL}), each client has a similar local objective due to the IID data and homogenous devices. 
As a result, the global update follows a similar optimization direction as observed in centralized training (Fig. \ref{Centralized training}). 
Unfortunately, in practice, individual clients in FL often exhibit significant heterogeneity, 
as depicted in Fig. \ref{Heterogeneous FL}.  
First, \emph{Heterogeneous data} induces diverse local optimization directions, 
straying from global minima \cite{scaffold}. 
Moreover, \emph{heterogeneous devices} (e.g., client 3 in Fig. \ref{Heterogeneous FL}) with varying computational speeds and communication capacities exacerbate the situation \cite{FedProx}, 
leading to a decreased convergence rate and compromised performance. 
Recently, corresponding solutions have been proposed to address the non-IID issue from different FL steps, 
including variance reduction \cite{scaffold}, regularization \cite{FedProx, FedDyn, FedDECORR}, contrastive learning \cite{MOON}, and sharpness-aware optimization \cite{FedSAM, FedSAM2} during local training, 
as well as normalized averaging \cite{FedNova}, layer-wise aggregation \cite{FedMA, FedLA}, momentum update \cite{FedAvgM, Slowmo, FedOpt}, and outlier pruning \cite{Outlier} during server aggregation. 
Additionally, personalized federated learning (pFL) methods \cite{FedAMP, FedLA} tackle the non-IID issue by training a customized local model on each client.
However, these methods are unaware of gradient conflicts across clients or investigate the relationship between the non-IID issue and the gradient conflict phenomenon in FL.
On the other hand, PCGrad \cite{PCGrad} first identifies that conflicting gradients lead to performance degradation in multi-task learning. 
Subsequently, \cite{DGGS} extend this insight to domain generation, presenting two gradient agreement strategies to alleviate gradient conflict. 
In contrast, we center on the gradient conflict phenomenon during server aggregation in FL caused by the non-IID issue. 

In this letter, we propose FedGH, a simple yet effective method that addresses local drifts through Gradient Harmonization. 
Since the non-IID issue causes weight divergence \cite{divergence}, we first study the gradient conflict phenomenon that emerges between local models on the server side. 
Intriguingly, we find that as the degree of non-IIDness increases, 
the severity of gradient conflicts escalates. Motivated by this observation, 
we subsequently explore the mitigation of local drifts through gradient deconfliction. 
Specifically, if two gradients conflict, we execute a projection of one gradient onto the orthogonal plane of the other. 
This gradient harmonization strategy aims to enhance client consensus during server aggregation. 
Finally, we conduct extensive experiments on four widely used benchmarks, including CIFAR-10, CIFAR-100, Tiny-ImageNet, and LEAF \cite{LEAF}, with diverse degrees of non-IID challenge. 
Experimental results demonstrate that FedGH yields more performance boosting in more severe heterogeneous scenarios while maintaining a consistent improvement over multiple baselines \cite{FedAvg, FedProx, FedNova, FedDECORR}, 
including the latest state-of-the-art FedDecorr \cite{FedDECORR}. 
As a plug-and-play module, FedGH can easily integrate with any FL framework, requiring no hyperparameter tuning.

Our contributions can be summarized as follows:

(1) We uncover that non-IID issue induces gradient conflicts between local models in FL, where stronger heterogeneity
leads to more severe gradient conflicts.

(2) We propose a novel method called FedGH, which effectively mitigates local drifts during server aggregation via gradient harmonization. 

(3) We validate the efficacy of FedGH through comprehensive experiments, showcasing its consistent improvement over multiple baselines.

\label{sec:intro}

%% file: sections/3_methodology.tex
\section{Proposed Method}

\subsection{Problem Definition}
FedGH aims to collaboratively learn a generalized model $w$ from $K$ clients over the dataset $ \mathcal{D} = \cup_{k \in [K]} \mathcal{D}_k $. 
The global objective is min$_w L(w) = \sum_{k=1}^{K} \frac{n_k}{n} L_k(w)$, 
where $n = \sum_{k=1}^{K} n_k$ is the number of samples in $\mathcal{D}$. For client $k$, $L_k(w) = \mathbb{E}_{(x, y) \sim \mathcal{D}_k} \ell(w; (x, y))$, 
where $\mathcal{D}_k$ is the local dataset.

\vspace{-0.2cm}
\subsection{Gradient Conflict in Heterogeneous FL}\label{Methodology}

We first probe the gradient conflict phenomenon under different degrees of non-IIDness. 
Toward this end, we simulate the non-IID data among different clients using the Dirichlet distribution \cite{FedAvgM, MOON, FedDECORR}. 
Specifically, we sample $P_c \sim Dir_K(\alpha)$ and allocate a $p_c^k \in P_c$ proportion of the samples of class $c$ to client $k \in [K]$. 
Here, $\alpha$ denotes the concentration parameter, with smaller values indicating stronger heterogeneity. 
Then, we set $\alpha \in \{0.01, 0.1\}$ and implement FedAvg \cite{FedAvg} to train a global convolutional neural network (CNN) utilized in \cite{FedOpt} on the CIFAR-10 dataset. 
Assuming that each client has uploaded the local gradient $g_t$ at the $t$-th communication round, where gradients from distinct layers are flattened to a 1-d vector. 
We identify gradient conflict when the cosine similarity between $g_t^i$ and $g_t^j, \forall i\neq j$, becomes negative, 
i.e., $ \frac{g_t^i \cdot g_t^j}{\| {g}_{t}^i \| \cdot \| {g}_{t}^j \|}  < 0$. Note that we will omit the denominator to simplify the expression of the cosine similarity sign.

\begin{figure}
    \centering
    \subfloat[]{\includegraphics[width=0.18\textwidth]{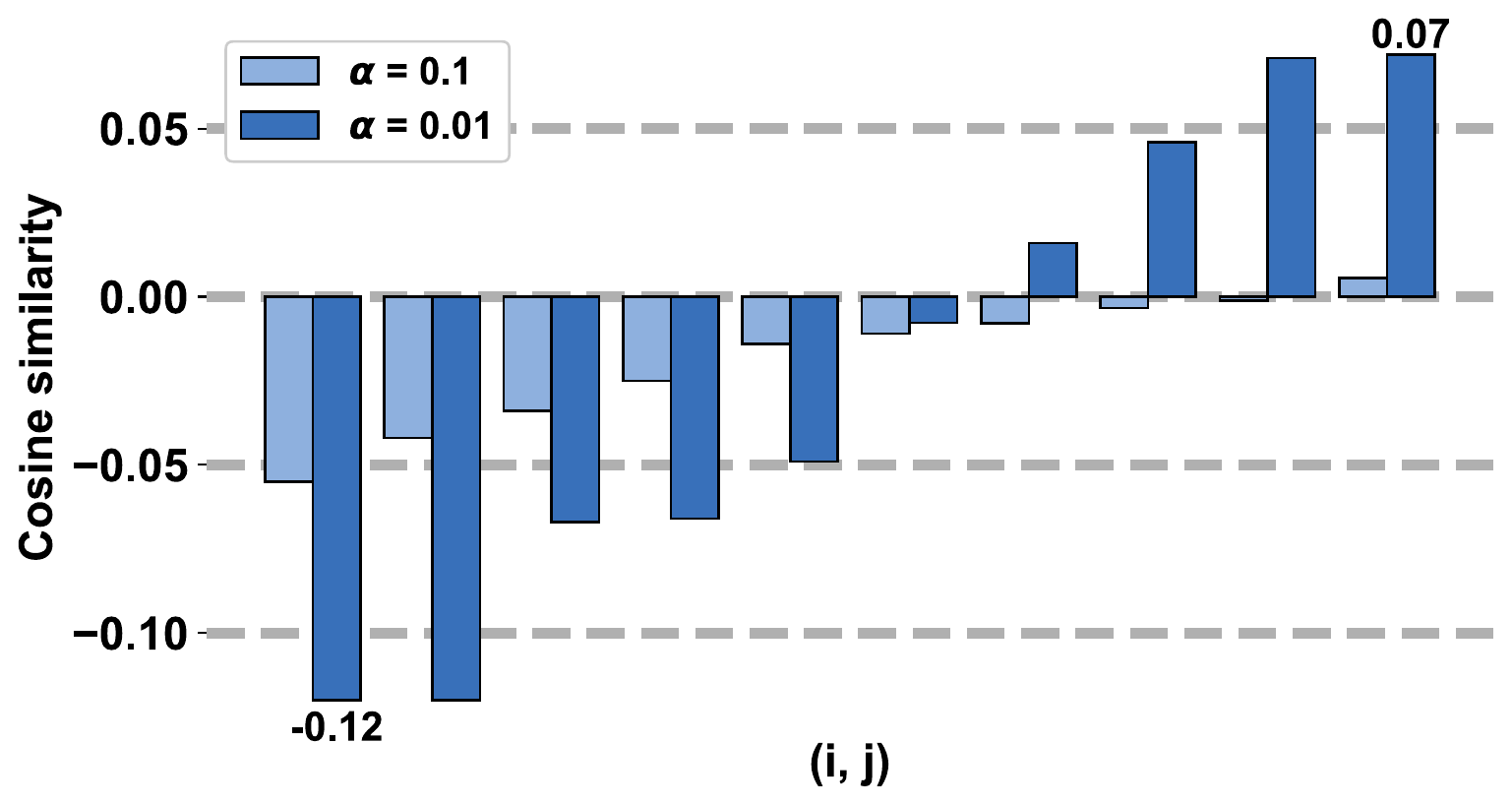}\label{gradient conflict1}} \hspace{0.4cm}
    \subfloat[]{\includegraphics[width=0.18\textwidth]{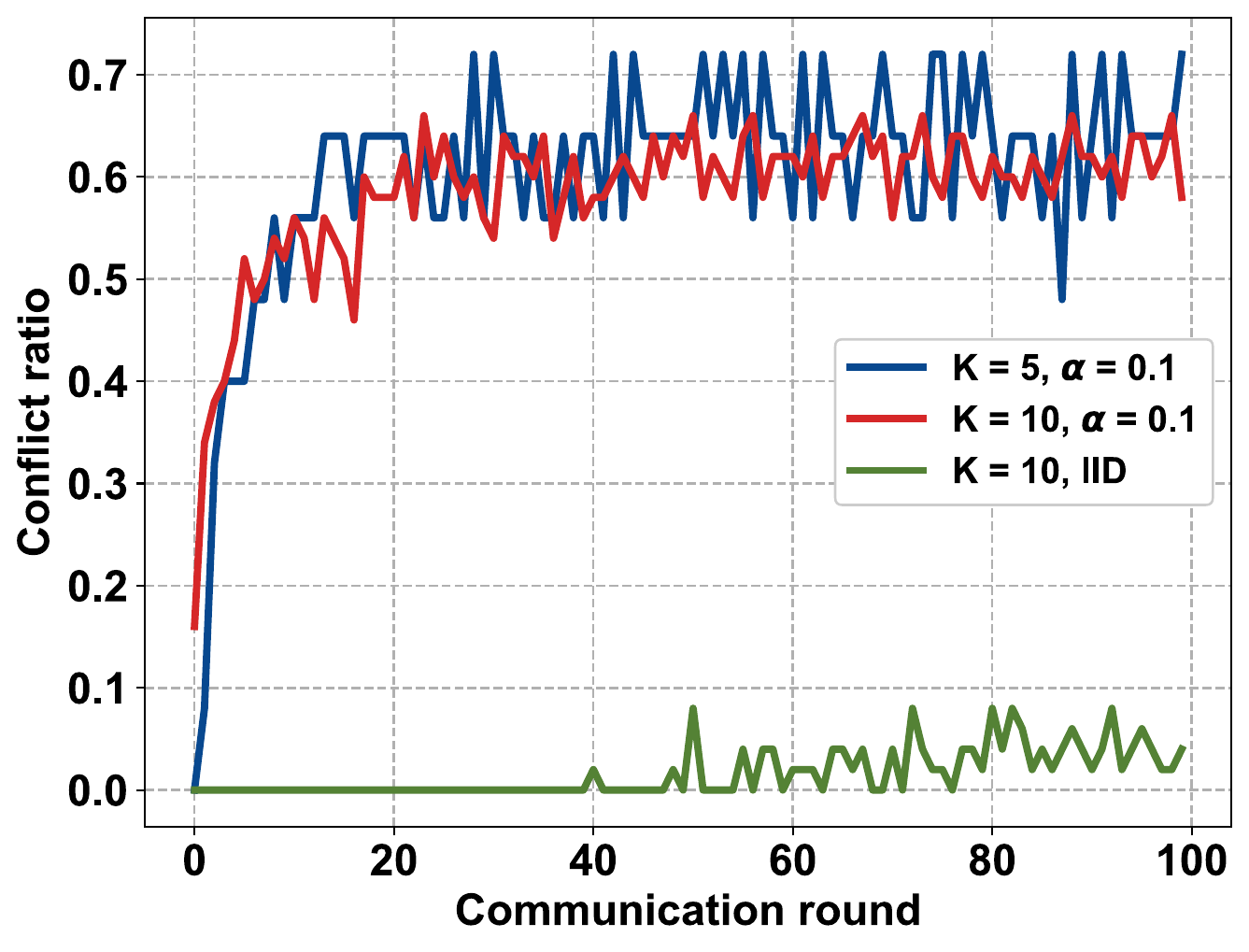}\label{gradient conflict3}} \\ \vspace{-0.4cm}
    \subfloat[]{\includegraphics[width=0.385\textwidth]{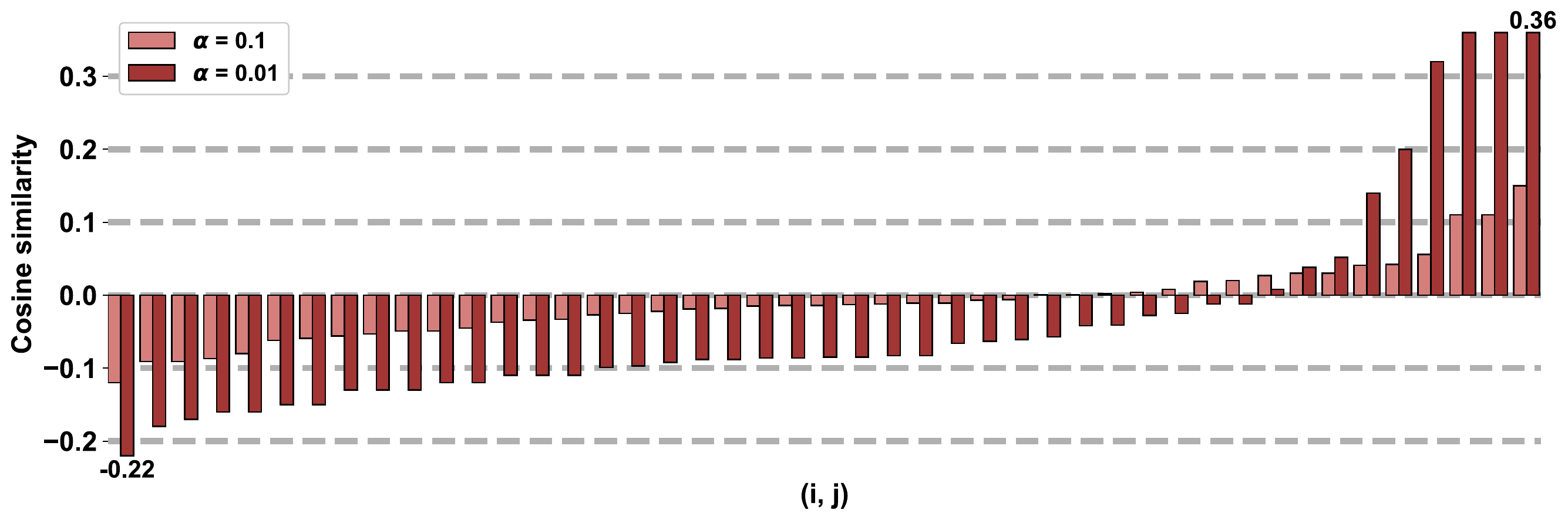}\label{gradient conflict2}}
    \caption{The non-IID issue causes gradient conflicts among (a) 5 and (c) 10 clients, with $\alpha$ = 0.1 and 0.01, respectively. The $x$-axis is client index pairs sorted by $y$, while the $y$-axis
    shows cosine similarity between gradient vectors. The change in gradient conflict ratio during training is depicted in (b).}
    \label{gradient conflict}
    \vspace{-15pt}
\end{figure}

First, we run 100 communication rounds involving 5 clients within two distinct non-IID scenarios. 
Subsequently, we visualize the sorted gradient similarity between each pair of clients in the final round, 
as illustrated in Fig. \ref{gradient conflict1}. Our observation reveals the presence of gradient conflicts in heterogeneous settings, 
while a stronger heterogeneity results in a more severe conflict, 
as evidenced by the minimum similarity decreasing from -0.06 to -0.12. 
Next, we increase the number of clients to 10; the results are presented in Fig. \ref{gradient conflict2}. 
We find that a larger number of clients further exacerbates the gradient conflict phenomenon, 
indicated by the reduction in minimum similarity from -0.12 to -0.22. 
Finally, we observe from Fig. \ref{gradient conflict3} that gradient conflicts become more frequent as the training advances. 
This phenomenon is consistent with the understanding that the non-IID issue causes local minima to deviate from global minima \cite{scaffold}. 
Moreover, gradient conflict is widespread in FL, even within the IID setting, 
and becomes notably more severe in non-IID scenarios. 
These experiments demonstrate that stronger heterogeneity leads to more severe local drifts, 
consequently inducing more violent gradient conflicts.
\vspace{-0.2cm}
\subsection{Tackling the Non-IID Issue by Gradient Harmonization}

Motivated by intuitive insights on gradient conflicts in FL, we propose FedGH, a simple yet effective method, 
to tackle the non-IID issue by gradient harmonization. FedGH proceeds as follows:
(1) FedGH first computes the cosine similarity between each gradient vector pair ($g_t^i$, $g_t^j$), $\forall i \neq j$. 
If the value is negative, indicating an obtuse angle between two gradient vectors, 
a gradient conflict between clients $i$ and $j$ is identified. (2) For each conflicting gradient pair ($g_t^i$, $g_t^j$), 
FedGH executes harmonization by projecting $g_t^i$ and $g_t^j$ onto each other's orthogonal planes as follows: 
\vspace{-0.2cm}
\begin{equation}
    \begin{aligned}
        g_{t}^i & \leftarrow g_{t}^i - \frac{g_{t}^i \cdot \widetilde{g}_{t}^j}{{\| \widetilde{g}_{t}^j \|}^2} \widetilde{g}_{t}^j \\
        g_{t}^j & \leftarrow g_{t}^j - \frac{g_{t}^j \cdot \widetilde{g}_{t}^i}{{\| \widetilde{g}_{t}^i \|}^2} \widetilde{g}_{t}^i 
    \end{aligned}
    \label{SD loss}
\end{equation}
\noindent Where $\widetilde G_t = (\widetilde g_t^1, \widetilde g_t^2, …, \widetilde g_t^K)$ is a duplicate of $G_t$.
Finally, the server aggregates the updated local gradients to acquire the global model. 
The complete FedGH procedure is detailed in Algorithm \ref{Pseudocode}.

\begin{algorithm}
    \SetAlgoNoEnd
    \setstretch{0.9}
    \caption{Federated Averaging with \textcolor{purple}{Gradient Harmonization (\texttt{FedGH})}}
    \label{Pseudocode}
    % \LinesNumbered
    \KwIn{number of communication rounds $T$, number of clients $K$, learning rate $\eta$,
    local minibatch size $B$, number of local epochs $E$}
    \KwOut{global model $w_{T}$}

    \SetKwProg{Fn}{}{:}{}
    \Fn {\rm \textbf{Server executes}}{
        initialize $w_0$ \\
        \For{each round $t = 0, ..., T-1$}{
            $C_t \leftarrow$ (random set of $C$ from $K$ clients) \\
            \For{client $ k \in C_t$ in parallel}{
                $g_{t+1}^k \leftarrow$ ClientUpdate($k$, $w_t$) \\
            }
            \textcolor{purple}{
            $\widetilde{G}_{t+1} \leftarrow G_{t+1}$ \\
            \For{$k \in C_t$}{
                \For{$j \in S_t \backslash k$ in random order}{
                    \If{$g_{t+1}^k \cdot \widetilde{g}_{t+1}^j < 0$}{
                        $g_{t+1}^k \leftarrow g_{t+1}^k - \frac{g_{t+1}^k \cdot \widetilde{g}_{t+1}^j}{{\| \widetilde{g}_{t+1}^j\|}^2} \widetilde{g}_{t+1}^j$
                    }
                }
            }
            }
            ${w}_{t+1} \leftarrow {w}_{t} + \sum_{k \in C_t} \frac{n_k}{n} g_{t+1}^k$ \\
        }
    \KwRet $w_{T}$
    }

    \SetKwProg{Fn}{}{:}{}
    \Fn {\rm \textbf{ClientUpdate}({$k$, $w_t$})}{
        $w_{t+1}^k \leftarrow w_t$ \\
        $\mathcal{B} \leftarrow $ (split $\mathcal{D}_k$ into batches of size $B$) \\
        \For{local epoch $i = 0, ..., E-1$}{
            \For{batch $b \in \mathcal{B}$}{
            $w_{t+1}^k \leftarrow w_{t+1}^k -\eta \nabla \ell (w_{t+1}^k; b)$
            }
        }
        \KwRet $w_{t+1}^k - w_t$
    }
\end{algorithm}

\begin{figure}
    \vspace{-0.2cm}
    \centering
    \subfloat[]{\includegraphics[width=0.235\textwidth]{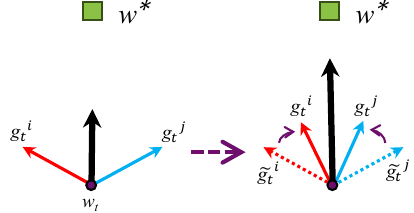}} \hspace{0.15cm}
    \subfloat[]{\includegraphics[width=0.235\textwidth]{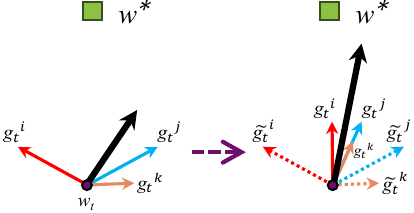}}
    \caption{Illustration of FedGH's effectiveness. If gradient conflict occurs between clients $i$ and $j$ in round $t$, FedGH projects $g_t^i$ and $g_t^j$ onto each other's orthogonal planes. We highlight that FedGH yields a faster convergence rate in (a) while effectively mitigating local drifts from global minima in (b).}
    \label{effectiveness}
    \vspace{-0.2cm}
\end{figure}

The effectiveness of FedGH is illustrated in Fig. \ref{effectiveness}.
 In a scenario where the sum of two conflicting clients exhibits an optimization direction towards global minima, 
 FedGH yields a faster convergence rate by discarding conflicting components of the gradient vector. 
 Conversely, in cases where the optimization direction deviates from global minima, 
 FedGH mitigates local drifts and rectifies the global update by enhancing local consensus across all clients. 
 In our extensive experiments, consistent performance boosting has been achieved on different benchmarks based on multiple baselines, 
 proving the significance of minimizing gradient conflicts in FL.

\label{sec:methodology}

%% file: sections/4_experiments.tex
\section{Experiments}
\subsection{Implementation Details}\label{Implementation}
We reproduce four competitive FL baselines, 
including FedAvg \cite{FedAvg}, FedProx \cite{FedProx}, FedNova \cite{FedNova}, and the state-of-the-art FedDecorr \cite{FedDECORR}.
We set $\mu$ and $\beta$ to 0.1 in FedProx and FedDecorr to achieve optimal performance, respectively. 

We conduct extensive experiments in four widely used benchmarks: 
CIFAR-10, CIFAR-100, Tiny-ImageNet, and LEAF \cite{LEAF}.   
For CIFAR-10, we use a CNN model with two convolution layers \cite{FedOpt}, as mentioned in Sec. \ref{Methodology}. 
For the more challenging CIFAR-100 and Tiny-ImageNet, we employ MobileNetV2 \cite{mobilenetv2}.
We partition the data into 20 clients with Autoaugment \cite{Autoaugment} and use the SGD optimizer with a momentum of 0.9. More settings are detailed in Tab. \ref{exp}.

LEAF is a comprehensive and challenging benchmark in FL that contains six tasks from computer vision (Celeba, FEMNIST), natural language processing (Twitter, Shakespeare, Reddit), and 
a synthetic classification task. We strictly follow the configuration of the data and model in \cite{LEAF} and slightly adjust some hyperparameters listed in Tab. \ref{exp} for better performance. 

\vspace{-0.3cm}
\begin{table}[H]
    \centering
    \setstretch{0.8}
    \caption{Details of the experimental environment, where $K$ is the number of clients, $C$ is the number of randomly selected clients in each round, $E$ is the number of local epochs, 
    $T$ is the number of communication rounds, $B$ is for batch size, and lr is for learning rate.}
    \label{exp}
    \vspace{-0.3em}
    \resizebox{1.0  \columnwidth}{!}{
    \setlength{\tabcolsep}{0.5mm}{}
    \begin{tabular}{llllll}
    \toprule
    Dataset       & Task                      & Model       & Non-IID setting  & FL setting           & Hyperparameters       \\ \midrule
    CIFAR-10      & Image Classification      & CNN         & Dirichlet($\alpha \in \{0.1, 1.0\})$  & $K$=20,$C \in \{4, 20\}$,$E$=5,$T$=100 & $B$=64,lr=0.01        \\
    CIFAR-100     & Image Classification      & CNN         & Dirichlet($\alpha \in \{0.01, 0.1\})$ & $K$=20,$C \in \{4, 20\}$,$E$=5,$T$=100 & $B$=64,lr=0.01      \\
    Tiny-ImageNet & Image Classification      & CNN         & Dirichlet($\alpha \in \{0.01, 0.1\})$ & $K$=20,$C \in \{4, 20\}$,$E$=5,$T$=50 & $B$=64,lr=0.01       \\ \midrule
    Celeba        & Image Classification      & CNN         & predefined in \cite{LEAF}  & $K$=467,$C$=50,$E$=1,$T$=100   & $B$=5,lr=0.01  \\
    FEMNIST       & Image Classification      & CNN         & predefined in \cite{LEAF}  & $K$=195,$C$=20,$E$=1,$T$=1000  & $B$=64,lr=0.01 \\
    Twitter       & Sentiment Analysis        & LSTM        & predefined in \cite{LEAF}  & $K$=132,$C$=10,$E$=1,$T$=100   & $B$=5,lr=0.01  \\
    Synthetic     & Classification            & Perceptron  & predefined in \cite{LEAF}  & $K$=1000,$C$=50,$E$=1,$T$=100  & $B$=5,lr=0.1   \\
    Shakespeare   & Next Character Prediction & LSTM        & predefined in \cite{LEAF}  & $K$=660,$C$=50,$E$=1,$T$=50    & $B$=64,lr=1.0  \\
    Reddit        & Language Modeling         & LSTM        & predefined in \cite{LEAF}  & $K$=813,$C$=50,$E$=1,$T$=100   & $B$=5,lr=1.0   \\ \bottomrule
    \end{tabular}
    }
\end{table}

To validate the effectiveness of FedGH, we simulate multiple non-IID scenarios from cross-silo and cross-device perspectives \cite{FedOpt}. 
First, we generate \emph{heterogeneous data} using the Dirichlet distribution detailed in Sec. \ref{Methodology} for CIFAR-10, CIFAR-100, and Tiny-ImageNet. 
As for LEAF, we follow the canonical non-IID sampling strategy in \cite{LEAF} to construct heterogeneous data.
Moreover, we simulate \emph{heterogeneous devices} by randomly selecting $C$ clients from $K$ clients in each communication round.
\vspace{-0.2cm}
\subsection{Performance Boosting}

\begin{table*}
    \centering
    \setstretch{0.8}
    \caption{Top-1 test accuracy across multiple non-IID benchmarks. All results are reported from the last communication round. The highest accuracy in each scenario is highlighted in \textbf{Bold}.}
    \label{acc}
    \vspace{-0.5em}
    \setstretch{0.5}
    \resizebox{2.0  \columnwidth}{!}{
    \setlength{\tabcolsep}{1.5mm}{} 
    \begin{tabular}{lccccccccccccccccccccc}
    \toprule
    \multicolumn{1}{l}{\multirow{3}{*}{Method}} & \multicolumn{3}{c}{CIFAR-10(\%)}  & \multicolumn{3}{c}{CIFAR-100(\%)} & \multicolumn{3}{c}{Tiny-ImageNet(\%)} & \multicolumn{2}{c}{Celeba(\%)}  & \multicolumn{2}{c}{FEMNIST(\%)} & \multicolumn{2}{c}{Twitter(\%)} & \multicolumn{2}{c}{Synthetic(\%)} & \multicolumn{2}{c}{Shakespeare(\%)} & \multicolumn{2}{c}{Reddit(\%)} \\ \cmidrule(lr){2-4} \cmidrule(lr){5-7} \cmidrule(lr){8-10} \cmidrule(lr){11-12} \cmidrule(lr){13-14} \cmidrule(lr){15-16} \cmidrule(lr){17-18} \cmidrule(lr){19-20} \cmidrule(lr){21-22} 
    & \begin{tabular}[c]{@{}c@{}}$\alpha$=0.1\\  $C$=20\end{tabular} & \begin{tabular}[c]{@{}c@{}}$\alpha$=1.0\\  $C$=4\end{tabular} & \begin{tabular}[c]{@{}c@{}}$\alpha$=1.0\\  $C$=20\end{tabular} & \begin{tabular}[c]{@{}c@{}}$\alpha$=0.01\\  $C$=20\end{tabular} & \begin{tabular}[c]{@{}c@{}}$\alpha$=0.1\\  $C$=4\end{tabular} & \begin{tabular}[c]{@{}c@{}}$\alpha$=0.1\\  $C$=20\end{tabular} & \begin{tabular}[c]{@{}c@{}}$\alpha$=0.01\\  $C$=20\end{tabular} & \begin{tabular}[c]{@{}c@{}}$\alpha$=0.1\\  $C$=4\end{tabular} & \begin{tabular}[c]{@{}c@{}}$\alpha$=0.1\\  $C$=20\end{tabular} & \begin{tabular}[c]{@{}c@{}}\multirow{1}{*}{$Acc_l$}\end{tabular} & \begin{tabular}[c]{@{}c@{}}\multirow{1}{*}{$Acc_g$}\end{tabular} & \begin{tabular}[c]{@{}c@{}}\multirow{1}{*}{$Acc_l$}\end{tabular} & \begin{tabular}[c]{@{}c@{}}\multirow{1}{*}{$Acc_g$}\end{tabular} & \begin{tabular}[c]{@{}c@{}}\multirow{1}{*}{$Acc_l$}\end{tabular} & \begin{tabular}[c]{@{}c@{}}\multirow{1}{*}{$Acc_g$}\end{tabular} 
    & \begin{tabular}[c]{@{}c@{}}\multirow{1}{*}{$Acc_l$}\end{tabular} & \begin{tabular}[c]{@{}c@{}}\multirow{1}{*}{$Acc_g$}\end{tabular} & \begin{tabular}[c]{@{}c@{}}\multirow{1}{*}{$Acc_l$}\end{tabular} & \begin{tabular}[c]{@{}c@{}}\multirow{1}{*}{$Acc_g$}\end{tabular} & \begin{tabular}[c]{@{}c@{}}\multirow{1}{*}{$Acc_l$}\end{tabular} & \begin{tabular}[c]{@{}c@{}}\multirow{1}{*}{$Acc_g$}\end{tabular} \\ \midrule
    FedAvg   & 68.66       & 65.40      & 75.28       & 41.00     & 42.15      & 61.50     & 31.44       & 25.98      & 46.64    & 75.67   & 86.25  & 79.46  & 78.43  & 63.54  & 64.51   & 84.37  & 78.38  & 47.52 & 50.41 & 9.22 & 12.28 \\
    \cellcolor[HTML]{DCDCDC}+FedGH  & \cellcolor[HTML]{DCDCDC}70.67          & \cellcolor[HTML]{DCDCDC}66.00      & \cellcolor[HTML]{DCDCDC}75.61   & \cellcolor[HTML]{DCDCDC}43.86      & \cellcolor[HTML]{DCDCDC}45.80        & \cellcolor[HTML]{DCDCDC}62.68    & \cellcolor[HTML]{DCDCDC}32.94    & \cellcolor[HTML]{DCDCDC}26.28    & \cellcolor[HTML]{DCDCDC}47.45    & \cellcolor[HTML]{DCDCDC}81.76 & \cellcolor[HTML]{DCDCDC}\textbf{89.69} & \cellcolor[HTML]{DCDCDC}\textbf{84.50}   & \cellcolor[HTML]{DCDCDC}81.85    & \cellcolor[HTML]{DCDCDC}66.72   & \cellcolor[HTML]{DCDCDC}\textbf{66.55}  & \cellcolor[HTML]{DCDCDC}85.10   & \cellcolor[HTML]{DCDCDC}82.45   & \cellcolor[HTML]{DCDCDC}47.74  & \cellcolor[HTML]{DCDCDC}50.79  & \cellcolor[HTML]{DCDCDC}\textbf{9.70}  & \cellcolor[HTML]{DCDCDC}12.48  \\ \midrule
    FedProx  & 69.63       & 65.32      & 74.84       & 41.05     & 42.38      & 61.46     & 31.42       & 25.91      & 46.67    & 78.62   & 86.86      & 79.35   & 78.66   & 63.39   & 64.45  & 84.74 & 78.38  & 47.58 & 50.54 & 9.48 & 12.21  \\
    \cellcolor[HTML]{DCDCDC}+FedGH     & \cellcolor[HTML]{DCDCDC}\textbf{71.80}  & \cellcolor[HTML]{DCDCDC}\textbf{66.27}      & \cellcolor[HTML]{DCDCDC}75.00    & \cellcolor[HTML]{DCDCDC}44.16      & \cellcolor[HTML]{DCDCDC}44.62        & \cellcolor[HTML]{DCDCDC}62.14    & \cellcolor[HTML]{DCDCDC}32.83       & \cellcolor[HTML]{DCDCDC}26.58      & \cellcolor[HTML]{DCDCDC}47.23  & \cellcolor[HTML]{DCDCDC}81.88  & \cellcolor[HTML]{DCDCDC}88.76  & \cellcolor[HTML]{DCDCDC}\textbf{84.50}    & \cellcolor[HTML]{DCDCDC}\textbf{81.99}   & \cellcolor[HTML]{DCDCDC}66.70   & \cellcolor[HTML]{DCDCDC}\textbf{66.55}  & \cellcolor[HTML]{DCDCDC}85.41   & \cellcolor[HTML]{DCDCDC}82.45   & \cellcolor[HTML]{DCDCDC}\textbf{48.17}  & \cellcolor[HTML]{DCDCDC}\textbf{50.91}  & \cellcolor[HTML]{DCDCDC}9.68  & \cellcolor[HTML]{DCDCDC}12.30  \\ \midrule
    FedNova  & 65.08       & 65.33      & 75.75       & 39.72     & 43.13      & 62.01     & 30.96       & 25.65      & 46.52    & 80.12   & 86.21      & 79.84   & 78.35   & 64.37   & 64.50  & 84.66  & 81.74  & 45.94 & 48.32 & 8.84 & 11.10 \\
    \cellcolor[HTML]{DCDCDC}+FedGH     & \cellcolor[HTML]{DCDCDC}67.56       & \cellcolor[HTML]{DCDCDC}66.15      & \cellcolor[HTML]{DCDCDC}\textbf{75.97}   & \cellcolor[HTML]{DCDCDC}42.75      & \cellcolor[HTML]{DCDCDC}44.66        & \cellcolor[HTML]{DCDCDC}\textbf{62.96}   & \cellcolor[HTML]{DCDCDC}33.03       & \cellcolor[HTML]{DCDCDC}26.74      & \cellcolor[HTML]{DCDCDC}47.13  & \cellcolor[HTML]{DCDCDC}81.21  & \cellcolor[HTML]{DCDCDC}89.09  & \cellcolor[HTML]{DCDCDC}83.33    & \cellcolor[HTML]{DCDCDC}81.12   & \cellcolor[HTML]{DCDCDC}\textbf{67.13}  & \cellcolor[HTML]{DCDCDC}66.26  & \cellcolor[HTML]{DCDCDC}86.02   & \cellcolor[HTML]{DCDCDC}\textbf{83.95}  & \cellcolor[HTML]{DCDCDC}46.56  & \cellcolor[HTML]{DCDCDC}49.04  & \cellcolor[HTML]{DCDCDC}9.14  & \cellcolor[HTML]{DCDCDC}11.55  \\ \midrule
    FedDecorr & 61.04       & 61.61      & 69.50       & 47.32     & 45.51      & 62.14    & 36.36       & 29.76      & 47.87     & 80.38     & 86.47    & 77.19   & 76.19   & 63.44   & 64.28  & 83.10 & 77.23  & 47.42 & 50.34 & 9.46 & 12.41 \\
    \cellcolor[HTML]{DCDCDC}+FedGH  & \cellcolor[HTML]{DCDCDC}60.61       & \cellcolor[HTML]{DCDCDC}62.71      & \cellcolor[HTML]{DCDCDC}70.33   & \cellcolor[HTML]{DCDCDC}\textbf{49.21}      & \cellcolor[HTML]{DCDCDC}\textbf{47.59}      & \cellcolor[HTML]{DCDCDC}62.88   & \cellcolor[HTML]{DCDCDC}\textbf{37.89}       & \cellcolor[HTML]{DCDCDC}\textbf{30.31}      & \cellcolor[HTML]{DCDCDC}\textbf{48.60}  & \cellcolor[HTML]{DCDCDC}\textbf{82.69}  & \cellcolor[HTML]{DCDCDC}88.77   & \cellcolor[HTML]{DCDCDC}81.54   & \cellcolor[HTML]{DCDCDC}80.31      & \cellcolor[HTML]{DCDCDC}66.56     & \cellcolor[HTML]{DCDCDC}65.22  & \cellcolor[HTML]{DCDCDC}\textbf{86.96}  & \cellcolor[HTML]{DCDCDC}82.11  & \cellcolor[HTML]{DCDCDC}47.78   & \cellcolor[HTML]{DCDCDC}50.47  & \cellcolor[HTML]{DCDCDC}\textbf{9.70}  & \cellcolor[HTML]{DCDCDC}\textbf{12.57}  \\ \bottomrule
    \end{tabular}
    }
\end{table*}

\begin{table*}
    \centering
    \setstretch{0.8}
    \vspace{-1em}
    \caption{Communication rounds and parallel time consumption to reach target top-1 accuracy. Here, $\alpha$=0.1 in CIFAR-10, and $\alpha$=0.01 in CIFAR-100 and Tiny-ImageNet.
    We use $Acc_g$ as the target accuracy in LEAF and $(\cdot \times)$ represents speedup brought by FedGH to the baseline. The most efficient results in each benchmark are highlighted in \textbf{Bold}.}
    \label{eff}
    \vspace{-0.5em}
    \setstretch{0.5}
    \resizebox{2.0  \columnwidth}{!}{
    \setlength{\tabcolsep}{0.3mm}{}
    \begin{tabular}{lcccccccccccccccccccc}
    \toprule
    \multicolumn{1}{l}{\multirow{2}{*}{Method}} & \multicolumn{2}{c}{CIFAR-10(60\%)} & \multicolumn{2}{c}{CIFAR-100(40\%)}  & \multicolumn{2}{c}{Tiny-ImageNet(30\%)}  & \multicolumn{2}{c}{Celeba(85\%)} & \multicolumn{2}{c}{FEMNIST(75\%)}  & \multicolumn{2}{c}{Twitter(60\%)}  & \multicolumn{2}{c}{Synthetic(80\%)}  & \multicolumn{2}{c}{Shakespeare(45\%)}  & \multicolumn{2}{c}{Reddit(10\%)} \\ \cmidrule(lr){2-3} \cmidrule(lr){4-5} \cmidrule(lr){6-7} \cmidrule(lr){8-9} \cmidrule(lr){10-11} \cmidrule(lr){12-13} \cmidrule(lr){14-15} \cmidrule(lr){16-17} \cmidrule(lr){18-19}
    \multicolumn{1}{c}{} & \multicolumn{1}{c}{Rounds} & \multicolumn{1}{c}{Time} & \multicolumn{1}{c}{Rounds} & \multicolumn{1}{c}{Time} & \multicolumn{1}{c}{Rounds} & \multicolumn{1}{c}{Time} & \multicolumn{1}{c}{Rounds} & \multicolumn{1}{c}{Time} & \multicolumn{1}{c}{Rounds} & \multicolumn{1}{c}{Time} & \multicolumn{1}{c}{Rounds} & \multicolumn{1}{c}{Time} & \multicolumn{1}{c}{Rounds} & \multicolumn{1}{c}{Time} & \multicolumn{1}{c}{Rounds} & \multicolumn{1}{c}{Time} & \multicolumn{1}{c}{Rounds} & \multicolumn{1}{c}{Time} \\ \midrule
    FedAvg      & 28  & 78  & 88  & 339 & 45 & 1447 & 82 & 35 & 784  & 26  & 51  & 23  & 68  & \textbf{7} & 13 & 12 & 35 & \textbf{5} \\
    \cellcolor[HTML]{DCDCDC}+FedGH      & \cellcolor[HTML]{DCDCDC}25(1.1$\times$)  & \cellcolor[HTML]{DCDCDC}70(1.1$\times$) & \cellcolor[HTML]{DCDCDC}72(1.2$\times$) & \cellcolor[HTML]{DCDCDC}290(1.2$\times$) & \cellcolor[HTML]{DCDCDC}39(1.2$\times$) & \cellcolor[HTML]{DCDCDC}1263(1.1$\times$)  & \cellcolor[HTML]{DCDCDC}44(1.9$\times$)  & \cellcolor[HTML]{DCDCDC}31(1.1$\times$) & \cellcolor[HTML]{DCDCDC}499(1.6$\times$) & \cellcolor[HTML]{DCDCDC}30(0.9$\times$) & \cellcolor[HTML]{DCDCDC}\textbf{39(1.3$\times$)} & \cellcolor[HTML]{DCDCDC}\textbf{19(1.2$\times$)}  & \cellcolor[HTML]{DCDCDC}56(1.2$\times$)  & \cellcolor[HTML]{DCDCDC}10(0.7$\times$)  & \cellcolor[HTML]{DCDCDC}13(1.0$\times$) & \cellcolor[HTML]{DCDCDC}14(0.9$\times$) & \cellcolor[HTML]{DCDCDC}25(1.4$\times$) & \cellcolor[HTML]{DCDCDC}7(0.7$\times$)  \\ \midrule
    FedProx     & 28  & 80  & 88  & 471 & 43 & 1413  & 68 & 29 & 749  & 24  & 55  & 28  & 68  & 8  & 12 & \textbf{11} & 35 & 6 \\
    \cellcolor[HTML]{DCDCDC}+FedGH      & \cellcolor[HTML]{DCDCDC}\textbf{21(1.3$\times$)}  & \cellcolor[HTML]{DCDCDC}\textbf{59(1.4$\times$)}  & \cellcolor[HTML]{DCDCDC}73(1.2$\times$)  & \cellcolor[HTML]{DCDCDC}407(1.2$\times$) & \cellcolor[HTML]{DCDCDC}39(1.1$\times$) & \cellcolor[HTML]{DCDCDC}1295(1.1$\times$)  & \cellcolor[HTML]{DCDCDC}41(1.7$\times$)  & \cellcolor[HTML]{DCDCDC}29(1.0$\times$) & \cellcolor[HTML]{DCDCDC}482(1.6$\times$)  & \cellcolor[HTML]{DCDCDC}31(0.8$\times$) & \cellcolor[HTML]{DCDCDC}\textbf{39(1.4$\times$)} & \cellcolor[HTML]{DCDCDC}21(1.3$\times$)  & \cellcolor[HTML]{DCDCDC}56(1.2$\times$)  & \cellcolor[HTML]{DCDCDC}12(0.7$\times$) & \cellcolor[HTML]{DCDCDC}\textbf{11(1.1$\times$)} & \cellcolor[HTML]{DCDCDC}14(0.8$\times$) & \cellcolor[HTML]{DCDCDC}25(1.4$\times$) & \cellcolor[HTML]{DCDCDC}8(0.8$\times$) \\ \midrule
    FedNova     & 50  & 244 & 90  & 584 & 46 & 1633  & 66 & 28 & 531  & \textbf{20}  & 55 & 27  & 53  & \textbf{7} & 39 & 40 & 68 & 10 \\
    \cellcolor[HTML]{DCDCDC}+FedGH      & \cellcolor[HTML]{DCDCDC}45(1.1$\times$)  & \cellcolor[HTML]{DCDCDC}235(1.0$\times$) & \cellcolor[HTML]{DCDCDC}71(1.3$\times$) & \cellcolor[HTML]{DCDCDC}424(1.4$\times$)  & \cellcolor[HTML]{DCDCDC}39(1.2$\times$) & \cellcolor[HTML]{DCDCDC}1397(1.2$\times$)  & \cellcolor[HTML]{DCDCDC}\textbf{29(2.3$\times$)}  & \cellcolor[HTML]{DCDCDC}\textbf{20(1.4$\times$)} & \cellcolor[HTML]{DCDCDC}\textbf{361(1.5$\times$)} & \cellcolor[HTML]{DCDCDC}22(0.9$\times$)  & \cellcolor[HTML]{DCDCDC}\textbf{39(1.4$\times$)} & \cellcolor[HTML]{DCDCDC}21(1.3$\times$)  & \cellcolor[HTML]{DCDCDC}\textbf{38(1.4$\times$)}  & \cellcolor[HTML]{DCDCDC}9(0.8$\times$) & \cellcolor[HTML]{DCDCDC}35(1.1$\times$) & \cellcolor[HTML]{DCDCDC}39(1.0$\times$) & \cellcolor[HTML]{DCDCDC}50(1.4$\times$) & \cellcolor[HTML]{DCDCDC}15(0.7$\times$)  \\ \midrule
    FedDecorr   & 63  & 424 & 52  & 284 & 29 & 863   & 73 & 32 & 900  & 29 & 60 & 33 & $>$100 & $>$10 & 12 & 24 & 30 & \textbf{5} \\
    \cellcolor[HTML]{DCDCDC}+FedGH      & \cellcolor[HTML]{DCDCDC}64(1.0$\times$)  & \cellcolor[HTML]{DCDCDC}443(1.0$\times$)  & \cellcolor[HTML]{DCDCDC}\textbf{43(1.2$\times$)} & \cellcolor[HTML]{DCDCDC}\textbf{243(1.2$\times$)} & \cellcolor[HTML]{DCDCDC}\textbf{27(1.1$\times$)} & \cellcolor[HTML]{DCDCDC}\textbf{807(1.1$\times$)}  & \cellcolor[HTML]{DCDCDC}45(1.6$\times$)  & \cellcolor[HTML]{DCDCDC}34(0.9$\times$)  & \cellcolor[HTML]{DCDCDC}621(1.4$\times$) & \cellcolor[HTML]{DCDCDC}33(0.9$\times$) & \cellcolor[HTML]{DCDCDC}56(1.1$\times$) & \cellcolor[HTML]{DCDCDC}33(1.0$\times$)  & \cellcolor[HTML]{DCDCDC}62  & \cellcolor[HTML]{DCDCDC}15 & \cellcolor[HTML]{DCDCDC}12(1.0$\times$) & \cellcolor[HTML]{DCDCDC}25(1.0$\times$) & \cellcolor[HTML]{DCDCDC}\textbf{23(1.3$\times$)} & \cellcolor[HTML]{DCDCDC}7(0.7$\times$)  \\ \bottomrule
    \end{tabular}
    }
    \vspace{-0.3cm}
\end{table*}

Tab. \ref{acc} demonstrates the top-1 test accuracy across multiple benchmarks under various non-IID scenarios. 
Noteworthy is that we fix the random seed for a fair comparison and report all results from the last communication round. For LEAF, 
we follow \cite{LEAF} to evaluate the mean accuracy of local clients denoted as $Acc_l$ and the global model performance $Acc_g$ used in other benchmarks. 
We can observe that FedGH consistently improves all FL baselines in diverse heterogeneous settings. 

Specifically, \textbf{(1) FedGH improves multiple baselines.} Each column in Tab. \ref{acc} shows the results of incorporating FedGH into different FL baselines, and FedGH consistently achieves substantial performance gains. 
For instance, in FEMNIST, FedGH enhances the performance of FedAvg and FedProx by 5.04\% and 5.15\%, respectively. Additionally, a notable improvement of 2.76\% is observed when FedGH is incorporated with FedNova in Twitter, resulting in a substantial 3.86\% enhancement when integrated with FedDecorr in Synthetic.
\textbf{(2) FedGH outperforms baselines across various tasks.} According to Tab. \ref{exp} and Tab. \ref{acc}, 
FedGH exhibits enhancements ranging from 1.53\% to 5.15\% in five image classification datasets. 
Moreover, in sentiment analysis, next character prediction, and language modeling tasks, FedGH improves up to 3.18\%, 0.59\%, and 0.48\%, respectively. 
Additionally, FedGH contributes to notable enhancements ranging from 0.67\% to 4.88\% in the synthetic classification task.
\textbf{(3) FedGH achieves greater performance enhancement in more severe non-IID scenarios.} 
We simulate three degrees of non-IIDness in CIFAR-10, CIFAR-100, and Tiny-ImageNet, where FedGH yields more substantial improvements in cases of stronger heterogeneity (i.e., smaller $\alpha$ and $C$).
As shown in Tab. \ref{acc}, this trend results in a 0.16\% to 1.18\% enhancement in less severe heterogeneity and a more pronounced 1.89\% to 3.11\% improvement in the strongest heterogeneous scenario. 
As elaborated in Section \ref{Methodology}, this effect can be attributed to stronger heterogeneity leading to more severe gradient conflicts between clients.

Furthermore, we evaluate the efficiency of FedGH. Tab. \ref{eff} presents the number of communication rounds and parallel time consumption ($C$ clients complete local updates simultaneously in each round) required to attain the target top-1 accuracy. 
Notably, FedGH demonstrates a substantial reduction in communication rounds across all benchmarks, achieving a remarkable 2.3$\times$ speedup in Celeba and a 1.6$\times$ speedup in FEMNIST. 
Regarding time consumption, despite the additional time cost associated with gradient projection in FedGH, the faster convergence rate leads to an overall decrease in total time consumption, 
particularly evident in CIFAR-10, CIFAR-100, Tiny-ImageNet, and Twitter. 
However, in datasets with fewer data per client, such as Synthetic and Reddit, FedGH introduces a relatively higher time overhead. 
It is crucial to highlight that FedGH enhances the final performance, as demonstrated in Tab. \ref{acc}. In summary, as a plug-and-play module, FedGH can be effectively and efficiently integrated into diverse FL frameworks \cite{FedAvg, FedDECORR, FedX}.

\begin{minipage}{0.5\textwidth}
    \vspace{-0.3cm}
    \hspace{-0.35cm}
    \begin{minipage}{0.45\textwidth}
        \begin{table}[H]
            \setlength{\tabcolsep}{1pt}
            \centering
            \setstretch{0.8}
            \caption{Ablation study on the number of cliens $K$.}
            \vspace{-0.5em}
            \setstretch{0.5}
            \label{clients}
            \begin{tabular}{clcc}
            \toprule
            $K   $ & Method & CIFAR-10 & CIFAR-100 \\ \midrule
            \multirow{2}{*}{5    }       & FedAvg        & 64.93        & 62.93           \\
                    & \cellcolor[HTML]{DCDCDC}+FedGH       & \cellcolor[HTML]{DCDCDC}64.75     & \cellcolor[HTML]{DCDCDC}63.14    \\ \midrule
            \multirow{2}{*}{10   }       & FedAvg       & 69.74        & 56.12           \\
                    & \cellcolor[HTML]{DCDCDC}+FedGH       & \cellcolor[HTML]{DCDCDC}72.23     & \cellcolor[HTML]{DCDCDC}56.84    \\ \midrule
            \multirow{2}{*}{20   }       & FedAvg       & 68.66        & 41.00           \\
                    & \cellcolor[HTML]{DCDCDC}+FedGH       & \cellcolor[HTML]{DCDCDC}70.67     & \cellcolor[HTML]{DCDCDC}43.86    \\ \midrule
            \multirow{2}{*}{30   }       & FedAvg       & 67.21        & 32.16           \\
                    & \cellcolor[HTML]{DCDCDC}+FedGH       & \cellcolor[HTML]{DCDCDC}67.04     & \cellcolor[HTML]{DCDCDC}36.21    \\ \bottomrule
                \end{tabular}
        \end{table}
    \end{minipage}
    \hspace{0cm}
    \begin{minipage}{0.45\textwidth}  
        \begin{table}[H]
            \setlength{\tabcolsep}{1pt}
            \centering
            \setstretch{0.8}
            \caption{Ablation study on the number of local epochs $E$.}
            \vspace{-0.5em}
            \setstretch{0.5}
            \label{epochs}
            \begin{tabular}{clcc}
            \toprule
            $E   $ & Method & CIFAR-10 & CIFAR-100 \\ \midrule
            \multirow{2}{*}{1   }        & FedAvg       & 57.45     & 29.49              \\
                    & \cellcolor[HTML]{DCDCDC}+FedGH       & \cellcolor[HTML]{DCDCDC}58.55     & \cellcolor[HTML]{DCDCDC}33.42     \\ \midrule
            \multirow{2}{*}{5   }        & FedAvg       & 68.66        & 41.00           \\
                    & \cellcolor[HTML]{DCDCDC}+FedGH       & \cellcolor[HTML]{DCDCDC}70.67     & \cellcolor[HTML]{DCDCDC}43.86     \\ \midrule
            \multirow{2}{*}{10   }        & FedAvg       & 69.56        & 45.01          \\
                    & \cellcolor[HTML]{DCDCDC}+FedGH       & \cellcolor[HTML]{DCDCDC}70.42     & \cellcolor[HTML]{DCDCDC}45.54     \\ \midrule
            \multirow{2}{*}{15   }        & FedAvg       & 67.86       & 44.33           \\
                    & \cellcolor[HTML]{DCDCDC}+FedGH       & \cellcolor[HTML]{DCDCDC}70.62     & \cellcolor[HTML]{DCDCDC}45.34     \\ \bottomrule
            \end{tabular}
        \end{table}
    \end{minipage}   
\end{minipage}

\subsection{Ablation study on the number of clients}
As discussed in Sec. \ref{Methodology}, having more clients increases the occurrences of gradient conflicts. 
Thus, we investigate the effectiveness of FedGH with different numbers of clients ($K$). 
We test FedGH in conjunction with FedAvg under the challenging non-IID scenario, 
where $\alpha$ is set to 0.1 and 0.01 for CIFAR-10 and CIFAR-100, respectively. 
As shown in Tab. \ref{clients}, it is evident that FedGH slightly impacts the baseline when $K$ = 5. 
However, FedGH significantly enhances FedAvg as $K$ increases, resulting in a 2.01\% to 2.49\% improvement in CIFAR-10 and 0.72\% to 4.05\% in CIFAR-100. 
This observation underscores that FedGH is particularly suitable for cross-device FL scenarios involving a large number of clients \cite{FedSAM}.

\vspace{-0.1cm}
\subsection{Ablation study on the number of local epochs}
We also conduct an ablation study on the number of local epochs ($E$) under the same non-IID scenario, 
as shown in Tab. \ref{epochs}. In CIFAR-10, we consistently observe performance improvements by FedGH from 0.86\% to 2.76\%. 
Additionally, when $E$ is set to 15, each client is susceptible to overfitting in heterogeneous data with a skewed distribution. 
However, FedGH maintains high performance compared to $E$ = 10, while FedAvg exhibits a notable degradation. 
This suggests that FedGH can mitigate local drifts and correct the optimization direction towards global minima. 
In CIFAR-100, FedGH achieves the most significant performance improvements of 3.93\% and 2.86\% at $E$ = 1 and $E$ = 5, 
respectively. These observations demonstrate that FedGH can expedite the convergence rate of the global model, especially in scenarios where local computing power is constrained \cite{constrained}. 

\label{sec:experiments}